\documentclass[letterpaper, 10 pt, conference]{ieeeconf}  

\IEEEoverridecommandlockouts 

\usepackage{graphicx}
\usepackage{glossaries}
\usepackage{amsfonts}
\usepackage{amsmath}
\usepackage{amssymb}
\usepackage{caption}
\usepackage{tabularx}
\usepackage{algorithm2e}
\usepackage{textcomp}
\usepackage{tikz}   
\usepackage{graphicx}
\usepackage{subfigure}
\usepackage{subcaption}

\usepackage{svg}
\usepackage{amsfonts}
\usepackage{siunitx}
\usepackage{amssymb}

\usepackage{algpseudocode}
\RestyleAlgo{ruled}

\usepackage{amsmath}

\newacronym{dof}{DoF}{Degrees of Freedom}
\newacronym{mri}{MRI}{Magnetic Resonance Imaging}

\begin{document}

\title{\LARGE \bf Design of Magnetic Continuum Robots with Tunable Force Response Using Rotational Ring Pairs}

\author{Alex Sayres, Giovanni~Pittiglio,~\IEEEmembership{Member,~IEEE}  
\thanks{FuTURE Lab, Department of Robotics Engineering, Worcester Polytechnic Insitute (WPI), Worcester, MA 01605, USA. Email: {\tt\small \{asayres, gpittiglio\}@wpi.edu}}}

\maketitle

\begin{abstract}
In this paper, we discuss a novel continuum robot design that enables the online tuning of the magnetic response at its tip. The proposed method allows for the change of both effective magnetic direction and intensity, introducing steering \glspl{dof} without the need to control the external fields. This is unattainable with classical designs, which rely on fixed internal magnetic content and steer solely under the effect of a controllable magnetic field. The proposed robot design can be used in both controllable and fixed magnetic fields, potentially widening the clinical applicability of these robots. We experimentally show a max tip deflection of 33.8 mm from the resting state (23\% of the length). We discuss a model based on modified beam theory that captures the mechanical behavior of the continuum robot, with a mean absolute tip tracking error of 1.86\:mm (1.2\% of the robot’s length) and maximum errors of less than 4.8\:mm (3.2\% of the length) for all experimental points.
\glsresetall
\end{abstract}


\IEEEpeerreviewmaketitle
\section{Introduction}
\label{sec:introduction}
Continuum robots with tip \cite{Edelmann2017MagneticDevices} and distributed magnetic elements \cite{Nelson2024RemoteTelesurgery, Pittiglio2024ContinuumChains, Kim2022TeleroboticManipulation, Pittiglio2023ClosedRobots} have brought significant improvements in both navigation tasks -- such as vascular catheterization \cite{Kim2022TeleroboticManipulation, Pancaldi2025Flow-drivenEmbolization} and endoscopy \cite{Pittiglio2022Patient-SpecificEndoscopy, Bruns2020MagneticallyResults} -- and contact stabilization, e.g., in cardiac procedures \cite{Pittiglio2024MagneticTreatment, Chautems2018DesignAblations} and endoscopic imaging \cite{Greenidge2025HarnessingImaging}. These robots' ability to steer inside the anatomy, powered by the interaction of on-board magnetic elements with external fields, makes them perfect candidates for applications that require accuracy and a dynamic response. 

Magnetic robots traditionally rely on the control of external sources of magnetic fields. While this enables miniaturization and front-end force control, generating controlled magnetic fields remains a challenge. Often, this requires controlling large movable magnets \cite{Carpi2009StereotaxisEndoscopy, Pittiglio2023CollaborativeMagnets, Kim2022TeleroboticManipulation} or powering energy-consuming systems of coils \cite{Edelmann2017MagneticDevices, Dreyfus2024DexterousAccess}.

Tuning the magnetic response of the robot itself, instead of that of the field generated by the external source, has been investigated for applications where the latter cannot be naturally tuned. This is often the case in \gls{mri} machines because the nature of the machine's imaging requires that the homogeneous magnetic field remain constant. In these cases, a few alternatives may allow controlled behavior in magnetic catheters, such as using an electromagnetic tip \cite{Phelan2022DesignNeuroendoscope} or controlling the re-magnetization of the tip magnets using strong fields \cite{Tiryaki2023MagneticFields}. 

Electromagnetic alternatives have the potential for more refined control of the catheter's tip; however, they are more challenging to fabricate at sub-millimeter scales. Moreover, their magnetic field density is more limited than that of permanent magnets, making them a better approach in very high magnetic fields, such as in \gls{mri} contexts. Similarly, the re-magnetization of ferromagnetic tips \cite{Tiryaki2023MagneticFields} is only possible in high fields above the magnet's coercitivity. This is $\sim$ \SI{1}{\tesla} for standard Neodymium magnets, while most non-\gls{mri} actuation systems operate in the milli-Tesla range.

\begin{figure}
    \centering
    \includegraphics[width=\columnwidth]{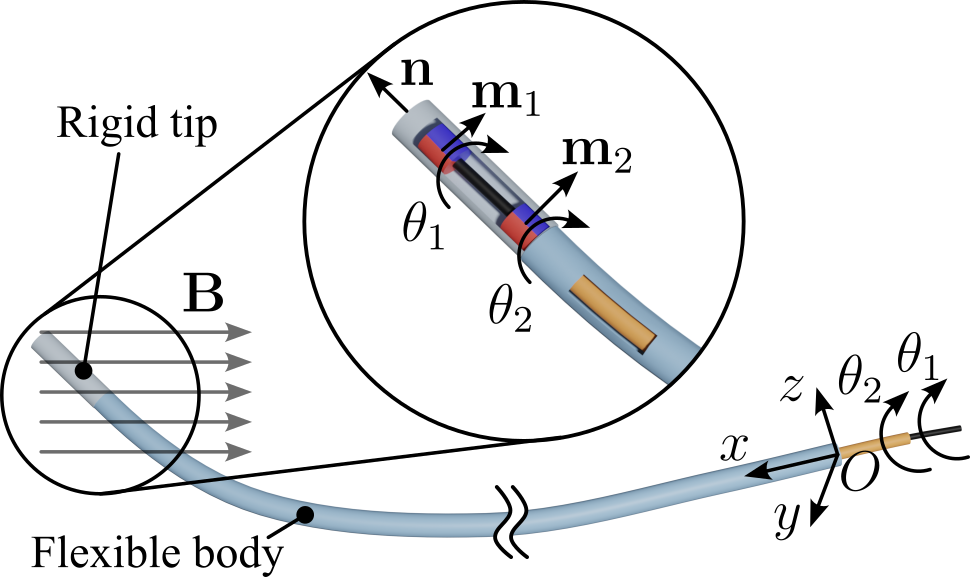}
    \caption{Magnetic continuum robot with tunable magnetic response.}
    \label{fig:main_fig}
\end{figure}

We propose an alternative approach that leverages mechanical tuning, i.e., modifying the configuration of the magnets within the continuum robot to obtain a tunable force/torque response at the tip. 
Figure \ref{fig:main_fig} illustrates the proposed design: the robot's tip contains two small magnets mounted inside a rigid casing, positioned at a fixed distance from each other along the robot’s central axis. A spacer may be used to keep their relative positions constant, while two flexible concentric tubes allow each magnet to rotate independently around the tip’s axis. The tubes are selected to be flexible in bending but resistant to twisting; thus, any rotation applied at their base is transmitted to the tip, enabling precise control of the magnets’ orientations.

To obtain a tip-only response, we rely on the relative rotation of the internal magnets while constraining their relative translation, thereby limiting shape effects previously studied by Park \emph{et al.} \cite{Park2024WorkspaceMagnet}. We employ straight, flexible coaxial tubes solely leveraging the robot's magnetic response while minimizing magneto-mechanical effects that occur with pre-bent concentric tubes \cite{Peyron2022MagneticAnalysis}. 

In this paper, we introduce the robot's design along with a model derived from modified beam theory (Section \ref{sec:overall_model}) and detail its experimental analysis (Section \ref{sec:experiments}). The model's mechanical parameters are trained using real-world data (Section \ref{sub:tuning}), and through experimental validation in 2D, we show that the tuned model achieves a \SI{4.8}{\milli\meter} maximum error -- 3.2\% of the robot's length (Section \ref{sub:validation}). We eventually demonstrate the design's ability to cover a 3D workspace through fine control of its effective tip magnetization in Section \ref{sub:evaluation}. Conclusions are presented in Section \ref{sec:conclusions}.

\section{Robot Model}
\label{sec:overall_model}
We present a modified Euler-Bernoulli model, where a generic source of magnetic field generates an external magnetic field to the continuum robot. We assume that the internal magnets can be approximated as dipoles, that the torsional stiffness of the continuum robot is infinite, and that the robot undergoes small tip deflection.

Figure \ref{fig:main_fig} shows the design of the continuum robot with a tunable magnetic tip. Two magnets with magnetic dipoles $\mathbf{m}_1\in\mathbb{R}^3$ and $\mathbf{m}_2 \in \mathbb{R}^3$, orthogonal to the rings' axes, are encased in the rigid tip of the robot at a distance $\delta \in \mathbb{R}$ along the tip tangent $\mathbf{n} \in \mathbb{R}^3$.

The magnets maintain at a constant relative position, while their individual angles of rotation around $\mathbf{n}$,  $\theta_1,\theta_2\in\mathbb{R}$, are controlled using two coaxial tubes. The tubes are flexible and are assumed to be infinitely rigid in torsion. Therefore, an axial rotation at their proximal end can be transferred to the tip, enabling the rotation of the two magnets.

\subsection{Magnetic Actuation Model}
\label{sec:design}
The magnetic force and torque on the $i$-th magnet subjected to an external field $\mathbf{B}(\mathbf p)\in\mathbb{R}^3$ are found as 
\begin{subequations}
\label{eq:force_torque_single}
\begin{align}
\mathbf{f}_i &= \nabla \left.\left(\mathbf{m}_i(\theta_i)\cdot\mathbf{B}(\mathbf p)\right)\right|_{\mathbf{p}=\mathbf{p}_i} = \left.\frac{\partial \mathbf{B}(\mathbf p)}{\partial \mathbf{p}}\right|_{\mathbf{p}=\mathbf{p}_i}^\top \mathbf{m}_i(\theta_i) \label{subeq:force_single}\\
\boldsymbol{\tau}_i &= \left.\left(\mathbf{m}_i(\theta_i) \times \mathbf{B}(\mathbf p)\right)\right|_{\mathbf{p}=\mathbf{p}_i} = -\left[\mathbf{B}(\mathbf p_i)\right]^\wedge \mathbf{m}_i(\theta_i) \label{subeq:torque_single}
\end{align}
\end{subequations}
with $\mathbf p_i \in \mathbb{R}^3$ the position of the $i$-th magnet; $\left[\mathbf{v}\right]^\wedge \triangleq \left(\mathbf{v} \times \mathbf{e}_1 \:|\: \mathbf{v} \times \mathbf{e}_2 \:|\: \mathbf{v} \times \mathbf{e}_3 \right) \in \mathbb{R}^{3\times3}$, $\mathbf{e}_j \in \mathbb{R}^3$ $j$-th element of the canonical basis of $\mathbb{R}^3$.

Because the magnets cannot move relative to each other and their rotation is mechanically controlled, we disregard their magnetic influence on one another. We can compute the overall magnetic forces and torque, considering only one external source of magnetic field as

\begin{subequations}
\label{eq:force_torque_total}
\begin{align}
\mathbf{f} &=  \sum_{i=1}^2\left.\frac{\partial \mathbf{B}(\mathbf p)}{\partial \mathbf p}\right|_{\mathbf{p}=\mathbf{p}_i}^\top \mathbf{m}_i(\theta_i) \label{subeq:force}\\
\boldsymbol{\tau} &= \sum_{i=1}^2-\left[\mathbf{B}(\mathbf p_i)\right]^\wedge \mathbf{m}_i(\theta_i) + \delta\left[\mathbf{n}\right]^\wedge \mathbf{f} =  \nonumber  \\
&\sum_{i=1}^2 \left(-\left[\mathbf{B}(\mathbf p_i)\right]^\wedge  + \delta\left[\mathbf{n}\right]^\wedge \left.\frac{\partial \mathbf{B}(\mathbf p)}{\partial \mathbf p}\right|_{\mathbf{p}=\mathbf{p}_i}^\top\right) \mathbf{m}_i(\theta_i) \label{subeq:torque}
\end{align}
\end{subequations}


The overall force and torque are controlled by the axial rotation of the magnets $\mathbf{q} = \left(\theta_1 \: \theta_2\right)^\top \in \mathbb{R}^2$.

\subsection{Continuum Robot Mechanics}
\label{sec:model}
We model the mechanical effects of the magnetic force and torque, $\mathbf{f}$ and $\boldsymbol{\tau}$ in (\ref{eq:force_torque_total}), to find the robot shape given a controlled configuration of $\mathbf{q}$. 

We assume that the continuum robot is fixed at the base, and the local tangent to its centerline is along the $x$ axis. Assuming the concentric tubes' axial rigidity dominates their bending stiffness, we approximate the robot as a cantilever beam with negligible elongation. We parameterize the robot by length $s\in\left[0, L\right]$ and assume that force $\mathbf{f}$ and moment $\boldsymbol{\tau}$ are applied at the tip $s=L$. The bending moment of the tube is 
\begin{equation}
    \mathbf{M}(s) = \boldsymbol{\tau} + (L - s)  \mathbf{e}_1 \times \mathbf{f} 
\end{equation}
with resulting curvature
\begin{equation}
    \boldsymbol{\kappa}(s) = \frac{1}{EI}\mathbf{M}(s) = \frac{1}{EI} \left(\boldsymbol{\tau} + (L - s)  \mathbf{e}_1 \times \mathbf{f} \right)
\end{equation}
where  $E$ is the elastic modulus, and $I$ is the second moment of area of the tube cross-section.
    

By assuming a small curvature, we find the tangent of the robot's centerline $\mathbf{t}(s)$ by integration 
\begin{equation}
        \mathbf{t}(s) \approx \mathbf{e}_1 + \int_0^s \boldsymbol{\kappa}(\xi) \times \mathbf{e}_1 \: \text{d}\xi .
        \label{eq:w_linears}
\end{equation}
The tip position $\mathbf{p} = \mathbf{p}_0 +  \int_0^{L}\mathbf{t}(s) \text{d}s$ and the tangent $\mathbf{n} \triangleq \mathbf{t}(L)$ are found as
\begin{subequations}
\label{eq:position_tangent}
\begin{align}
\mathbf{p} &= \mathbf{p}_0 + L\left(\mathbf{e}_1 + \frac{L}{EI}\left(\frac{1}{2}\boldsymbol{\tau} + \frac{L}{6} \mathbf{e}_1 \times \mathbf{f} \right)\times \mathbf{e}_1\right) \\
\mathbf{n} &= \mathbf{e}_1 + \frac{L}{EI}\left(\boldsymbol{\tau} + \frac{L}{2} \mathbf{e}_1 \times \mathbf{f} \right)\times \mathbf{e}_1
\end{align}
\end{subequations}
where $\mathbf{p}_0$ is the position of the robot's base. 
\label{section:mag_mech_model}


We use the vector triple product identity and the fact that the cross product is anti-commutative to rewrite (\ref{eq:position_tangent}) as 
\begin{subequations}
\label{eq:position_tangent_rewrite}
\begin{align}
\mathbf{p} &= \mathbf{p}_0 + L\left(\mathbf{e}_1 + \frac{L}{EI} \left(-\frac{1}{2}\left[\mathbf{e}_1\right]^\wedge\boldsymbol{\tau} + \frac{L}{6} (\mathbf{I} - \mathbf{e}_1 \mathbf{e}_1^\top) \mathbf{f} \right)\right) \\
\mathbf{n} &= \mathbf{e}_1 + \frac{L}{EI}\left(-\left[\mathbf{e}_1\right]^\wedge\boldsymbol{\tau} + \frac{L}{2} (\mathbf{I} - \mathbf{e}_1 \mathbf{e}_1^\top) \mathbf{f} \right)
\end{align}
\end{subequations}
and by grouping the terms such that $\mathbf{w} = \left(\mathbf{f}^\top | \boldsymbol{\tau}^\top\right)^\top$, we have
\begin{subequations}
\label{eq:position_tangent_group}
\begin{align}
\mathbf{p} &=  \mathbf{p}_0 + L\mathbf{e}_1 + \frac{L^2}{2EI} \left(\left. (\mathbf{I} - \mathbf{e}_1 \mathbf{e}_1^\top) \: \right| \: -\left[\mathbf{e}_1\right]^\wedge\right) \mathbf{w}  \\
\mathbf{n} &= \mathbf{e}_1 + \frac{L}{EI} \left(\left. (\mathbf{I} - \mathbf{e}_1 \mathbf{e}_1^\top) \: \right| \: -\left[\mathbf{e}_1\right]^\wedge\right) \mathbf{w}
\end{align}
\end{subequations}


By substituting (\ref{eq:force_torque_total}) into (\ref{eq:position_tangent_group}), we obtain the model for the continuum robot in Fig. \ref{fig:main_fig}. We show its accuracy by simulating the effects of an external magnetic field in Section \ref{sub:evaluation}, where we solve iteratively to find the steady-state tip pose $\left(\mathbf{p}, \mathbf{n}\right)$ for a specific input rotation of the ring magnets  ($\mathbf{q}$). 
For control purposes, one could define a desired set of $\left(\mathbf{p}, \mathbf{n}\right)$ and solve for $\mathbf{q}$.

\section{Experimental Validation}
\label{sec:experiments}
In the following, we report the experimental analysis of the model in Section \ref{sec:model}. We designed and tested a custom surgical catheter with internal rotating-tip magnets placed within a powerful magnetic field. We tracked the tip deflection and internal ring rotations using the setup shown in Fig. \ref{fig:setup} to compare the device's $\mathbf{p}(\mathbf{q)}$ to the model's predicted controls. 

\subsection{Experimental Setup}
\label{sub: Experimental Setup}
We built a demonstrator in which the continuum robot is made of a diametrically-magnetized N52 tip magnet (ID: \SI{2}{\milli\meter}, OD: \SI{4}{\milli\meter}, length: \SI{0.5}{\milli\meter}) connected to a super-elastic NiTi wire (\SI{0.4}{\milli\meter} diameter). A second equivalent magnet is mounted at the tip of a PTFE catheter (ID: \SI{1.2}{\milli\meter}, OD: \SI{1.8}{\milli\meter}, length: \SI{150}{\milli\meter}). Each magnet is mounted to its respective wire using a custom mounting piece. These mounting pieces constrain both magnets to the same axis of rotation and fix the axial translation of both magnets (assuming negligible axial strain). For this demonstrator, we found no need for a physical spacer, which allows for negligible inter-magnetic distance ($\delta \approx 0$), shown to provide the most symmetric workspace reachable by the robot’s tip. The interaction between the internal magnets could be easily counteracted by turning the coaxial elements of the robot. The mounting pieces also impose relative rotational limits on the magnets, allowing us to ``lock" the magnets in a high-potential-energy state by twisting the tube/wire enough so that any torsional strain is compensated.

The dimensions of the ring magnets were chosen to be the minimum size with a strong enough magnetic moment to be effectively deflected by the magnetic field of our experimental setup. In a more powerful magnetic field, ring magnets of smaller diameter could be used instead; thus further reducing the torsional effects between the internal magnets. The PTFE catheter's exceptionally low bending stiffness and reasonable torque transmission make it a good candidate for the outer catheter. Similarly, the NiTi wire's hyper-elasticity and low bending stiffness make it an effective choice.

\begin{figure}
    \centering
    \includegraphics[width=\columnwidth]{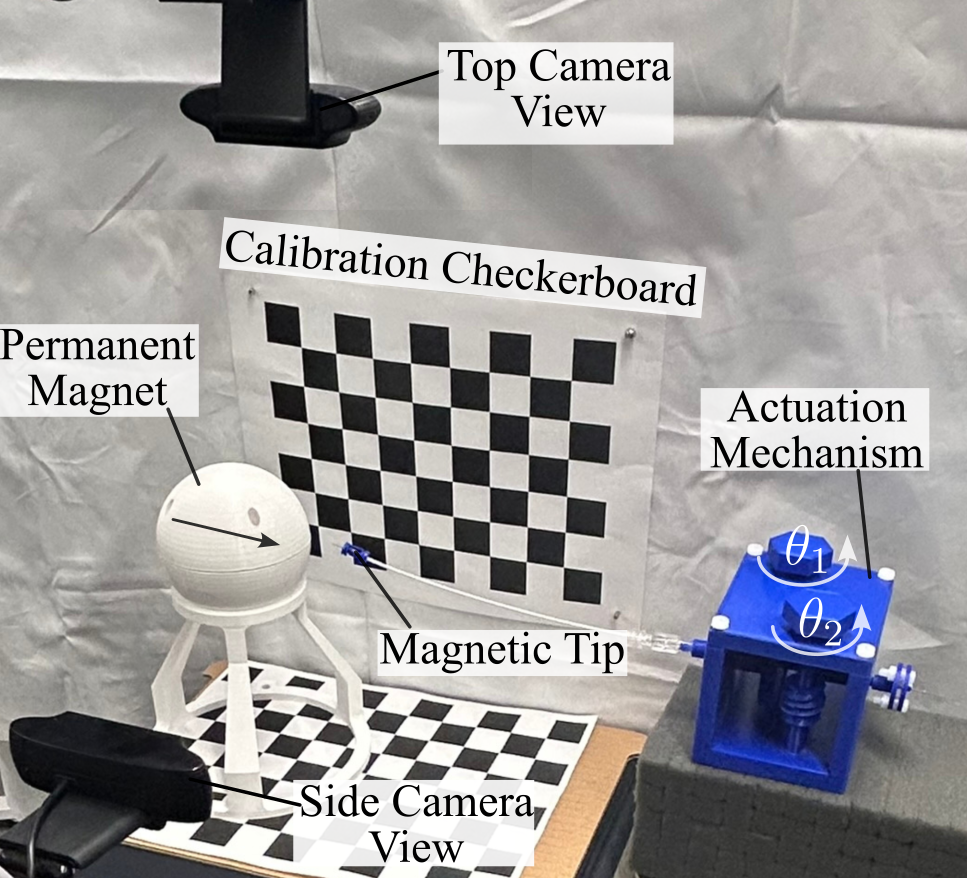}
    \caption{Setup for 2D and 3D tip tracking experiments.}
    \label{fig:setup}
\end{figure}

The NiTi wire runs through the PTFE catheter and actuates the distal magnet. The hyper-elastic nature of how the NiTi wire deforms ensures that the catheter returns to its pre-deflection position after the forces and torques are removed. Both the NiTi wire and the PTFE tube are connected to a manual actuation mechanism (Fig. \ref{fig:setup}) that can control each magnet's axial rotation independently. Both the PTFE catheter and the NiTi wire enter the hollow core of the demonstrator's gear shaft. The PTFE catheter is adhered via cyanoacrylate to the frontal spur gear, while the NiTi wire is adhered to a mounting piece in the back of the device. The rear external mount transmits torque from the rear spur gear while also allowing adjustable spacing of the magnets at the tip by axially shifting the mount and thus changing the length of the wire. The mechanism uses self-locking worm gears to drive the spur gears, allowing for precise rotational control of the tip while minimizing backlash. 

An external N52 cylindrical magnet (\SI{76.2}{\milli\meter} in diameter, \SI{38.1}{\milli\meter} in length) is positioned \SI{80}{\milli\meter} from the catheter's tip in the rest configuration. The external magnet's rotation is fixed, with its magnetic direction (shown by the arrow in Fig. \ref{fig:setup}) opposite to the catheter direction (-$x$ axis). The magnet is not moved or rotated during the experiments. 

The field properties of the magnet used in this demonstrator also influenced our decision to minimize the spacer distance between the magnets. In a more homogeneous field, each magnet would experience a similar magnetic potential from the external field, even if they were separated by some spacer distance. However, in our experimental setup using a (non-homogeneous) external field from a permanent cylindrical magnet, even small changes in the position of an internal magnet can impact the force and torque applied to it. If the magnets had a spacer distance, one magnet would always be closer to the external magnet, which would cause its rotation to more dominantly affect the steering of the catheter. This could make our demonstrator more difficult to control (especially keeping it in the neutral position). 

A camera is mounted on a UR10e 6DoF robotic arm and repositioned to obtain top and side camera views, ensuring its alignment with the top and side planes, respectively (Fig. \ref{fig:setup}). The same camera is moved between views during the experiments. Each camera view is calibrated to find the position of the catheter's tip in \SI{}{\milli\meter} in both planes. Markings along the tip of the catheter allow for tracking the rotation of the two magnets at the tip, accounting for possible mechanical loss between the actuation mechanism and the actual magnet orientation. The orientation is used as a measure of the tip magnets' angles $\hat \theta_1$ and $\hat \theta_2$.  

From each image, we extract the position of the tip on the plane $\mathbf{p}_y$ and the actual orientation of the distal magnet. 
To determine the axial rotations $\hat \theta_1$ and $\hat \theta_2$ from the planar images, we use a helical groove pattern in the mounting pieces of both magnets (similar to a screw, but with only one widely spaced thread). From a planar view, the groove appears as a notch in the material. The position of this notch follows a linear relationship with the rotation of the piece. Using the known dimensions of the tip, we can record the position of the notch and apply a linear transformation to find the rotation of the tip pieces. Figure \ref{fig:planar_pics}b reports the experimental results.

Using the setup in Fig. \ref{fig:setup}, we fixed the rotation of the proximal magnet $\theta_2 = 0$ and captured 16 pictures using the top camera view while adjusting the distal magnet by rotation of the actuation mechanism: $\theta_1 = \left[0, \: 12, \dots, \: 180\right]^\circ$. From the captured images, we extract the tip configuration of the robot; four configurations (including the minimum and maximum deflections) are shown in Fig. \ref{fig:planar_pics}a. Quantitatively, the demonstrator achieved a maximum deflection of \SI{33.8}{\milli\meter} (23\% of the length).

\subsection{Experimental Model Structure}
Using the relationships outlined in Sections \ref{sec:design} and \ref{sec:model}, we create a tunable model for the magnet's steady-state position as $\theta_1$ and $\theta_2$ change. The external magnetic field in our experiment is generated by a large permanent magnet, so the $\mathbf{B}(\mathbf{p})$ field is described as a dipole, which is generally a realistic approximation for the control of magnetic catheters \cite{Pittiglio2023CollaborativeMagnets}. The dipole model describes the field as 
\begin{equation}
    \mathbf{B}(\mathbf{p}) = \frac{{\mu_0}}{4\mathbf{\pi}}\left(\frac{\mathbf{3\mathbf{P}}(\mathbf{\mathbf{P}}\cdot {\mathbf{m}_e})-{\mathbf{m}_e}}{\mathbf{\mathbf{P}^3}}\right)
    \label{eq:dipole_field}
\end{equation}
where $\mathbf{m}_e$ is the dipole approximation of the external permanent magnet, and $\mathbf{P}$ is the position of the tip of the robot with respect to the center of the permanent magnet. We write $\mathbf{P} = \mathbf{p} - \mathbf{p}_e$; $\mathbf{p}_e$ is referred to as the magnet position in the robot's base frame, and $\mathbf{p}$ is the position of the robot tip from its base, as noted in (\ref{eq:position_tangent}). 

By substituting the definition of $\mathbf{B}(\mathbf{p})$ into (\ref{eq:force_torque_total}), we find the resultant force and moment at the tip of the robot and the relationship between the control parameters $\mathbf{q}$, the tip position, and the bending angle. 

To solve for the robot's tip position, we implemented an iterative solver in Matlab that considers the model nonlinearities arising from  (\ref{eq:position_tangent_group}) and (\ref{eq:dipole_field}) with initial guess is $\mathbf{p}^{(0)} = 150\mathbf{e}_1$, i.e., the robot's straight configuration.



\subsection{Model Tuning}
\label{sub:tuning}
Since the exact nominal elastic modulus $E$ could not be retrieved for the medical PTFE catheter, we use the elastic parameter as an unknown and tune it using the collected data. From observation, we find an initial value of the compound elastic modulus of $E\approx766$\SI{}{\mega\pascal}.
\begin{figure}[t]
    \centering
    \includegraphics[width=\columnwidth]{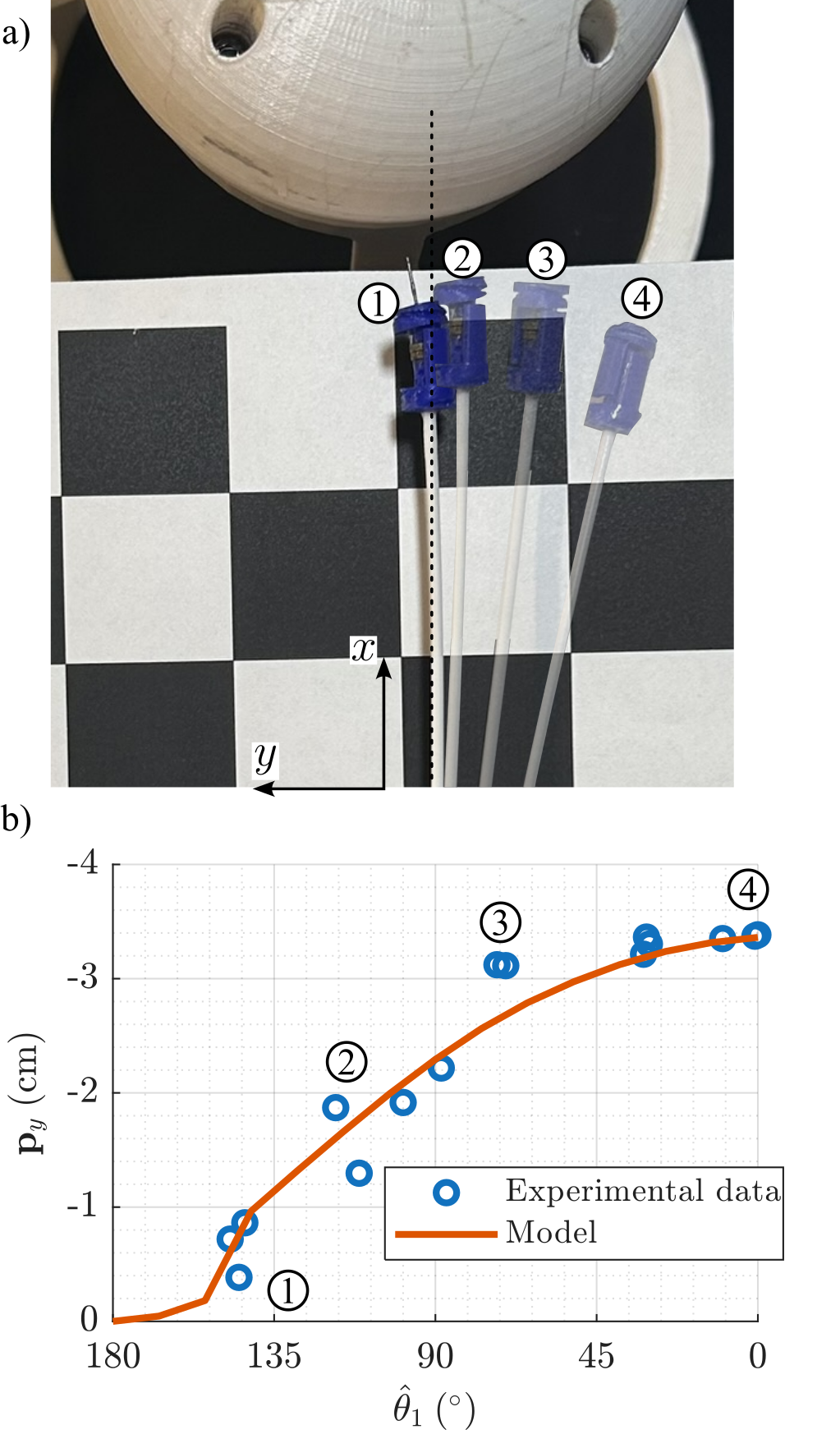}
    \caption{Experimental analysis of planar motion with constant $\theta_2=0$; a) top camera view; b) comparison of collected data and model.}
    \label{fig:planar_pics}
\end{figure}

\begin{figure}[t]
    \centering
    \includegraphics[width=\columnwidth]{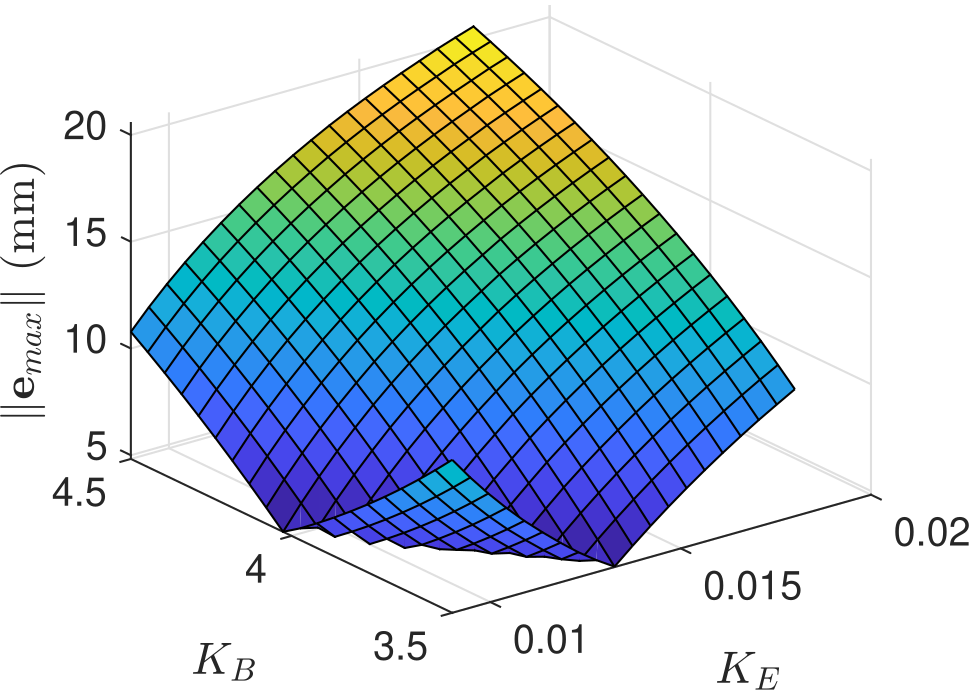}
    \caption{Maximum absolute error between model prediction and data for parametric calibration.}
    \label{fig:model_tuning}
\end{figure}

In the case of the magnetic response of the catheter, it did not accurately match the nominal model in (\ref{eq:dipole_field}), compared to other studies where internal magnetic elements were located further away from one another, e.g., \cite{Pittiglio2023ClosedRobots}, and when the external magnetic field was generated farther from the catheter's tip \cite{Pittiglio2023CollaborativeMagnets}. The effect of the latter is a larger discrepancy between the expected magnetic fields and gradients closer to the sources, which is mitigated further away \cite{Petruska2013OptimalApproximation}.

To account for the model-to-real-world discrepancy, we investigated different parametric modifications to the models in Sections \ref{sec:design} and \ref{sec:model}. We found an effective method using a linear scaling of the elastic modulus and scaling of the external magnet's field intensity and position.


To mitigate the field/gradient discrepancies observed in the effective magnetic field generated by the external magnetic field source, we use $K_B$ to adjust the nominal magnetic dipole intensity $\mathbf{m}_e' = K_B\mathbf{m}_e$ and the external magnet position $\mathbf{p}_e' = K_B \mathbf{p}_e$.  $K_B$ serves to create a more accurate balance of force and torque from the external field. 

We modify the model in (\ref{eq:dipole_field}) to account for the realistic magnetic field as 
\begin{equation}
    \mathbf{B}(\mathbf{p}) = \frac{K_B{\mu_0}}{4\mathbf{\pi}}\left(\frac{\mathbf{3\hat{\mathbf{P}}'}(\mathbf{\hat{\mathbf{P}}'}\cdot {\mathbf{m}_e})-{\mathbf{m}_e}}{\mathbf{\mathbf{P}'^3}}\right)
    \label{eq:dipole_field_KB}
\end{equation}
where $\mathbf{P}'= \mathbf{p} - K_B \mathbf{p}_e$.

The experimental data collected is represented in Fig. \ref{fig:planar_pics}. To find the model that best matches the data, we ran different iterations of our model while adjusting the parameters $K_E = \left[0.009, 0.018\right]$ and $K_B = \left[3.5, 4.5\right]$, for each iteration, we computed the error between the model-predicted tip position and the one measured in Fig. \ref{fig:planar_pics}: $\mathbf{e}_{max}$. The errors obtained across the parameter space are shown in Fig. \ref{fig:model_tuning}.

An exhaustive search is performed over the model-data comparison to find the minimum maximum error, i.e.
\begin{equation}
    K^*_E,K^*_B = \underset{K_E,K_B}{\arg\min} \: \mathbf{e}_{max}
\end{equation}
using the recorded data. The optimal values are found to be $K^*_B = 4.03$ and $K_E^* = 0.009$. While these coefficients unrealistically shrink our elastic modulus, the extremely low value of $K_E^*$ counters any net reduction to the field strength by $K_B^*$ that occurred when balancing the gradient. It is likely that different magnetic field conditions and model parameterizations would lead to more physically grounded results and will be further investigated in future work.

\subsection{Model Validation}
\label{sub:validation}
\begin{figure}[t]
    \centering
    \includegraphics[width=\columnwidth]{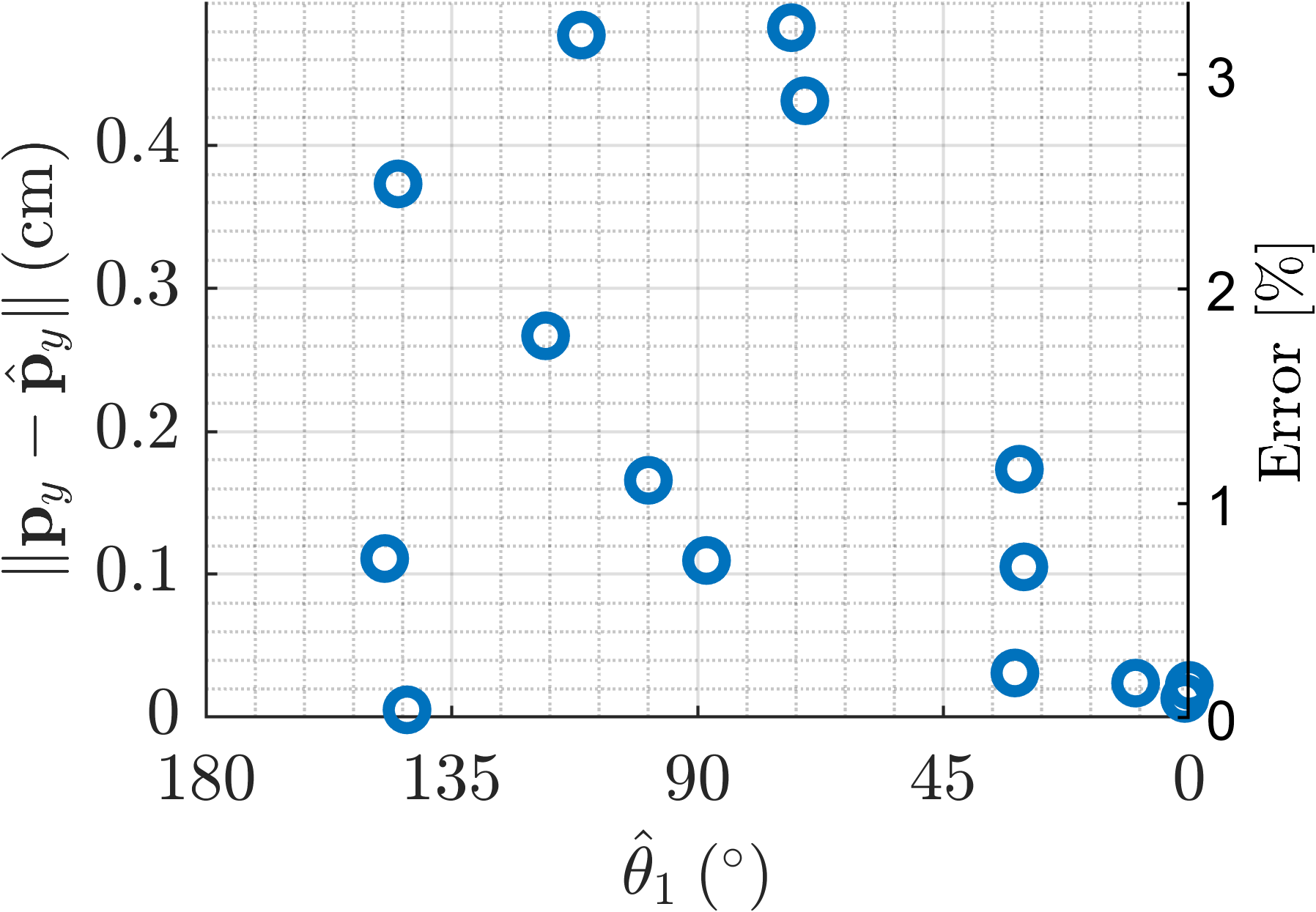}
    \caption{Absolute error between measured ($\mathbf{p}_y$) and predicted ($\hat{\mathbf{p}}_y$) position, reported in \SI{}{cm} and \% with respect to the robot's total length.}
    \label{fig:model_error}
\end{figure}

The error between the experimental data and the tuned model prediction is reported in Fig. \ref{fig:model_error}. When $\hat \theta_1$ is near its rotational limits such that the magnets are close to parallel or close to antiparallel, the error in our model is low (less than \SI{2.7}{\milli\meter})  -- approximately 1.8\% of the robots' total length. We notice the largest errors in the range $\hat \theta_1=[72,\: 111]$\SI{}{\degree} with a maximum error of \SI{4.8}{\milli\meter} -- 3.2\% of the robot's length. 

Due to the inherent nonlinearities of planar tracking an object floating in free space, we estimate that our camera tracking system has a maximum error of up to  $\pm $ \SI{1}{\milli\meter}. This measurement error is magnified for the tip rotational tracking measurement because the total range of the tracking notch is \SI{16}{\milli\meter} meaning that a $\pm $ \SI{1}{\milli\meter} planar measurement error of the notch corresponds to a $\pm $ \SI{8.2}{\degree} rotational measurement error. Accounting for the experimental measurement uncertainty, 12 of 16 data points are consistent with the model within the associated experimental error range. 

The 4 data points that fall outside of the experimental uncertainty (around \SI{70}{\degree}  and \SI{110}{\degree}) indicate that the deflection of the tip is influenced by factors our model does not yet account for. Firstly, we believe that the mechanical energy stored between the inner wire and the outer tube, combined with the inter-magnetic forces, causes some instabilities in these intermediary configurations. These small energy disturbances may be adding a chaotic element to the magnet's position that our model cannot account for.  The observation that the highest error data points all over-correct to low-energy magnetic resting states supports the idea that mechanical issues may primarily cause this error. This source of error is not linked to the model itself but rather its assumption of infinite torsional stiffness.  


A second source of error is the approximation of the internal magnets as perfect dipoles in air. The reality is that both small magnets may influence each other's effective magnetic response when in proximity to one another. 

Despite these potential discrepancies, we find that our calibrated model has an $R^2$ value of 0.946, indicating a solid fit across the domain of $\theta_1$.  The mean absolute error is found to be $\sim$\SI{1.86}{\milli\meter} (1.24\%), with a maximum error of \SI{4.8}{\milli\meter} (3.2\%), and a standard deviation of the error of \SI{1.9}{mm}. These error bounds are a promising indicator that the mechanism could be used to perform several endoscopic procedures. Based on these initial results, we will further improve the design and models to achieve sub-millimeter accuracy.

\subsection{Workspace Evaluation}
\label{sub:evaluation}
\begin{figure}
    \centering
    \includegraphics[width=\columnwidth]{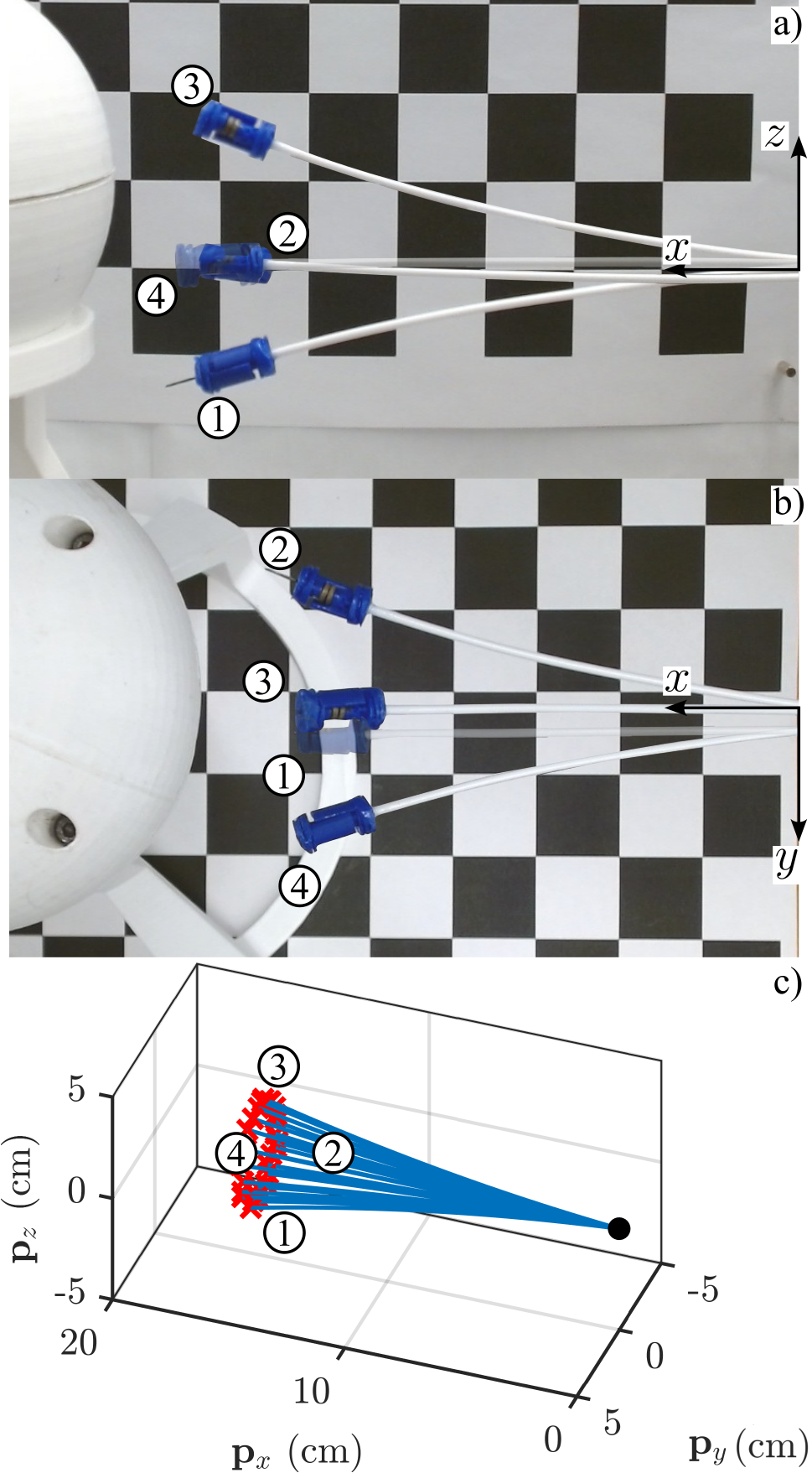}
    \caption{Workspace analysis in 3D; a) side camera view; b) top camera view; c) 3D reconstruction from bi-planar tracking.}
    \label{fig:pics_3D}
\end{figure}

To characterize the robot's workspace, we use the setup in Fig. \ref{fig:setup}, where both top and side camera views (Fig. \ref{fig:pics_3D}a and b) are used to estimate the position of the robot's tip on the $x-z$ and $x-y$ planes, respectively. The data from the bi-plane tracking system is merged to estimate the position of the robots' tip $\mathbf{p}$ in 3D, as represented in Fig. \ref{fig:pics_3D}c. During the experiments, both magnets' axial rotations $\theta_1$ and $\theta_2$ were manually actuated to obtain an elliptical pattern with the largest attainable bending.

Along the principal axes of motion, we see a maximum deflection of \SI{2.6}{\centi\meter} along $y$ and \SI{3.0}{\centi\meter} along $z$, indicating minimal differences in the amount of bending we can obtain along each axis -- approximately \SI{0.4}{\centi\meter}.

To study the homogeneity of the points, we interpolate the recorded tip positions in Fig. \ref{fig:pics_3D}c using an ellipse. We compute the distance between each experimental point and its closest point on the ellipse and obtain a maximum RMS error of \SI{0.44}{\centi\meter}, i.e., $\sim$18\% of the average deflection of \SI{2.45}{\centi\meter}. 

These studies indicate that the designed robot can cover an elliptical workspace $\sim$\SI{2.45}{\centi\meter} and that all points are found on the desired elliptical workspace contour, without singularities in any direction. In Section \ref{sub:validation} have shown that the robot's tip can be controlled to cover a continuous workspace on a plane, hence we can expect that -- with appropriate axial control of each magnet -- we can reach any point within the workspace. However, more testing is needed to quantitatively validate model accuracy across the entire workspace.

\section{Conclusions}
\label{sec:conclusions}
In this paper, we present a novel design for magnetic continuum robots that leverages coupled diametrically magnetized permanent magnets to tune the response to applied magnetic fields. The embedded magnets are independently actuated using torsionally rigid slender elements (catheters/wires) running through the robot to transmit rotational motion, effectively regulating the response to applied magnetic fields.

Under a constant applied magnetic field, we show a max tip deflection of \SI{33.8}{\milli\meter} from the resting state, i.e., 23\% of the total length.
A model is proposed that assumes the transmission elements are infinitely stiff in torsion and extension, that the robot's tip undergoes small deflections, and that the magnets are pure dipoles. This model's physics are tuned to real-world data, resulting in tip tracking errors of less than  $<$\SI{4.8}{\milli\meter}, or 3.2\% of the robot's length for all experimental points. Our model's mean absolute error is  \SI{1.86}{\milli\meter} or 1.24\% of the robot's length. Our results demonstrate the feasibility of this new design, which is capable of catheter steering without active control of the external field

In the future, we will investigate the effects of elements' non-negligible torsional effects and designs such as notched or composite materials that would allow such an assumption to hold in longer catheters than those used in our experimental tests. We will improve the torque response to obtain larger deflections and, consequently, investigate models such as the Cosserat-rod to capture larger deflections. We will continue to explore the modeling of magnetic components and examine a more suitable approximation than the dipole model for the two internal magnets in close proximity, potentially using higher order approximations \cite{Petruska2013OptimalApproximation}. We will also investigate how the model's accuracy can be improved when working in a more powerful homogeneous field.


\bibliographystyle{IEEEtran}
\bibliography{references}											
\end{document}